\pdfoutput=1

\documentclass[11pt]{article}

\usepackage[preprint]{acl}

\usepackage{times}
\usepackage{latexsym}

\usepackage[T1]{fontenc}

\usepackage[utf8]{inputenc}

\usepackage{microtype}

\usepackage{inconsolata}
\usepackage{hyperref}
\usepackage{url}
\usepackage{microtype}
\usepackage{booktabs}
\usepackage{tabularx}
\usepackage{multirow}
\usepackage{multicol}
\usepackage{algorithm}
\usepackage{algpseudocode}
\usepackage{float}
\usepackage{amsmath}
\usepackage{amsfonts}
\usepackage{verbatim}
\usepackage{graphicx}
\usepackage{relsize}
\usepackage{url}
\usepackage{xspace}
\usepackage{paralist}
\usepackage{cleveref}
\usepackage{adjustbox}
\usepackage{wrapfig}
\usepackage{subcaption}

\usepackage{ragged2e} 
\usepackage{makecell}

\usepackage{xcolor}
\usepackage{tcolorbox}
\definecolor{mycolor}{HTML}{2650CC}
\definecolor{highlight}{HTML}{81CE6D}

\newcommand{\ex}[1]{\textit{#1}\xspace}

\newcommand{\tabref}[1]{Table~\ref{#1}\xspace}
\newcommand{\figref}[1]{Figure~\ref{#1}\xspace}

\newcommand{\secref}[1]{Section\xspace\ref{#1}\xspace}

\title{Learning From Failure: Integrating Negative Examples when Fine-tuning \\ Large Language Models as Agents}

\author{Renxi Wang\textsuperscript{1,2} \qquad Haonan Li\textsuperscript{1,2} \qquad Xudong Han\textsuperscript{1,2} \\ 
\textbf{Yixuan Zhang\textsuperscript{1,2}} \qquad \textbf{Timothy Baldwin\textsuperscript{1,2,3}} \\
\textsuperscript{1}LibrAI\quad \quad \quad \quad \textsuperscript{2}MBZUAI\quad \quad \quad \quad
\textsuperscript{3}The University of Melbourne \\
  \texttt{\{renxi.wang,haonan.li,xudong.han,yixuan.zhang,timothy.baldwin\}@mbzuai.ac.ae}}

\begin{document}
\maketitle
\begin{abstract}

Large language models (LLMs) have achieved success in acting as agents, which interact with environments through tools such as search engines. However, LLMs are optimized for language generation instead of tool use during training or alignment, limiting their effectiveness as agents. To resolve this problem, previous work has first collected interaction trajectories between LLMs and environments, using only trajectories that successfully finished the task to fine-tune smaller models, making fine-tuning data scarce and acquiring it both difficult and costly. Discarding failed trajectories also leads to significant wastage of data and resources and limits the possible optimization paths during fine-tuning. 
In this paper, we argue that unsuccessful trajectories offer valuable insights, and LLMs can learn from these trajectories through appropriate quality control and fine-tuning strategies. By simply adding a prefix or suffix that tells the model whether to generate a successful trajectory during training, we improve model performance by a large margin on mathematical reasoning, multi-hop question answering, and strategic question answering tasks. We further analyze the inference results and find that our method provides a better trade-off between valuable information and errors in unsuccessful trajectories. To our knowledge, we are the first to demonstrate the value of negative trajectories and their application in agent-tunning scenarios. Our findings offer guidance for developing better agent-tuning methods and low-resource data usage techniques.
\footnote{Code and data are available at: \url{https://github.com/Reason-Wang/NAT}.}
\end{abstract}

\section{Introduction}
An agent is a model that has the ability to interact with environments, make decisions, and achieve predefined goals \citep{wooldridge1999intelligent}.
Early work used rule-based or template-based systems to complete tasks in narrow and specialized domains \citep{green1961baseball, weizenbaum1966eliza}.
Recent work has built off powerful LLMs such as GPT-4 \citep{achiam2023gpt}, using them as the core of an agent system to process information and make decisions \citep{Significant_Gravitas_AutoGPT, Yoheinakajima2023Babyagi}. 
This line of work has resulted in agent systems that are able to perform much more complex and general tasks.

However, these agents generally rely on closed-source LLMs through paid APIs, raising concerns about cost, latency, and reproducibility. Additionally, existing LLMs were not developed for agent use cases (e.g., generating actions or calling tools), and few-shot prompting offers only limited learning support \citep{Chen2023FireActTL}. 

Subsequent work has explored fine-tuning LLMs as agents, typically in three stages: data collection, fine-tuning, and inference \citep{Chen2023FireActTL,Zeng2023AgentTuningEG,Yin2023LumosLA,Qiao2024AUTOACTAA}. 
At the data collection stage, a powerful LLM such as GPT-4 is employed to interact with the environment, and the LLM-generated outputs and environment observations are collected as trajectories. In the fine-tuning stage, smaller models are fine-tuned using only successful trajectories. The fine-tuned models then serve as the agent's core during inference, demonstrating enhanced tool-using and decision-making capabilities, sometimes even surpassing the performance of the original LLM \citep{Yin2023LumosLA}.

To ensure the agent is being optimized appropriately, previous work has simply discarded trajectories that do not successfully complete the task (i.e.\ negative examples), using only successful trajectories (i.e.\ positive examples) in the fine-tuning stage \citep{Zeng2023AgentTuningEG,Chen2023FireActTL,Qiao2024AUTOACTAA}.  
However, in tasks demanding intricate planning, reasoning, or tool usage, the volume of discarded negative samples can exceed 60\%, leading to substantial data and computational resource wastage.

In this paper, we explore two key questions: (1) Can LLMs learn from these negative examples through fine-tuning? and (2) How can we optimize the use of negative examples to enhance agent performance? To address the first question, we fine-tune LLMs with a mix of positive and negative examples, and observe that incorporating negative examples generally yields benefits. 
For the second question, we introduce a negative-aware training (NAT) paradigm that explicitly tells the model to differentiate between correct and incorrect interactions by adding prefixes or suffixes. Our experiments demonstrate that NAT outperforms traditional methods by solely using positive examples or naively combining positive and negative ones, enabling better fine-tuning for low-resource data.
In addition, we conduct extensive experiments to analyze the learned agents' behavior after fine-tuning with negative examples.
Our contributions can be summarized as follows:
\begin{compactitem}
    \item We demonstrate the value of negative trajectories and introduce a negative-aware training paradigm, allowing LLM-based trained agents to effectively learn from both positive and negative examples. To the best of our knowledge, we are the first to utilize negative examples in agent training.
    \item We validate the broad applicability and effectiveness of learning from negative examples, and show that NAT enables models to acquire information akin to positive examples across various tasks and prompting strategies.
    \item We find NAT works by providing a better trade-off between useful information and errors in negative examples.
\end{compactitem}

\begin{figure*}[t]
    \centering
    \tiny
    \includegraphics[width=1.0\linewidth]{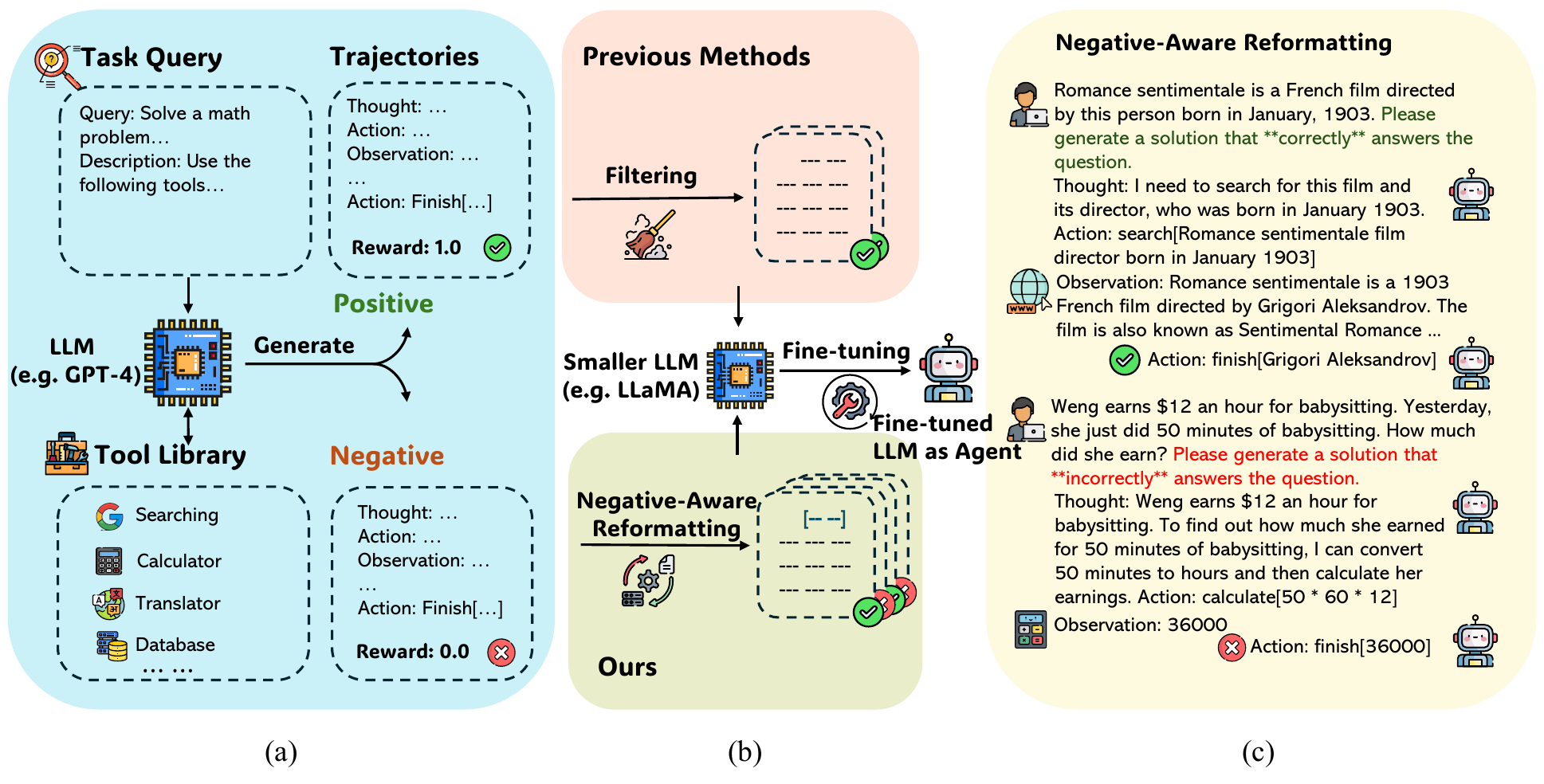}
    \caption{An overview of previous methods and our NAT paradigm. (a) Data collection, where interactions between LLMs and environments (tools) are collected. (b) Data processing, where previous methods simply filter out negative examples, while we reformat trajectories by adding prompts to task queries based on whether they are positive or negative. (c) An example of reformated positive and negative trajectories. We omit the system prompts here.}
    \label{fig:illustration}
\end{figure*}


\section{Related Work}
\subsection{Fine-tuning LLMs as Agents}
Previous work on language agents has taken a powerful LLM as the core of the agent system without fine-tuning \cite{sumers2023cognitive,wu2023autogen,Ruan2023IdentifyingTR,Zhao2023ExpeLLA}. However, LLMs are optimized to generate natural language. To make them capable of using tools and making decisions, current work typically collects trajectories generated by GPT-3.5/4, then uses these trajectories to fine-tune a smaller LLM \citep{Zeng2023AgentTuningEG,Chen2023FireActTL,Qiao2024AUTOACTAA,Chen2024AgentFLANDD,Zhang2024AgentOhanaDU,Zhou2024EnhancingTG}. \citet{Zeng2023AgentTuningEG} collect trajectories generated by GPT-4 on AgentBench \citep{Liu2023AgentBenchEL} tasks, and only keep samples that receive the best rewards. \citet{Chen2023FireActTL} collect trajectories on question answering tasks and fine-tune models with samples that correctly answer the question. \citet{Liu2024FromLT} propose a memory-enhanced agent framework and a complex filtering mechanism to collect fine-tuning datasets. \citet{Qiao2024AUTOACTAA} divide an agent into sub-agents with different functions. They then synthesize trajectories for the respective agents. However, they still only use samples with the best rewards. A simple ablation study was done by \citet{Zeng2023AgentTuningEG}. However, none of this work has investigated the effectiveness of negative samples in detail. 
Although not directly comparable, in \tabref{tab:others}, we provide the results of these methods and ours on several benchmarks for reference.

\subsection{Learning from Negative Results}

Learning from negative results can be divided into prompt-based and fine-tuning-based methods. Prompt-based methods enable LLMs to summarize experiences from previous mistakes without updating parameters. \citet{Madaan2023SelfRefineIR} use LLMs to first generate an output and then refine the output iteratively, while \citet{Shinn2023ReflexionLA} employ an evaluator to provide external feedback. \citet{Zhao2023ExpeLLA} let the agent compare successful and unsuccessful trajectories, and extract insights based on comparison. The success of these methods relies on the quality of the evaluator used to analyze the trajectories. The performance of fine-tuning-based methods is less predictable since model weights are updated, and less work has been done on this. \citet{Li2023TurningDI} propose a two-stage training paradigm to capture knowledge from negative samples. However, their method focuses on Chain-of-Thought prompts and is complex since multiple models are fine-tuned. Our work focuses on fine-tuning LLMs as agents and is much simpler and more effective.

\begin{table}[t]
\centering
\small
\resizebox{\linewidth}{!}{
\begin{tabular}{lccc}
\toprule
\textbf{Model} & \textbf{GSM8K} & \textbf{SVAMP} & \textbf{HotpotQA} \\
\midrule
AutoAct-7B  & -- & -- & 29.2 \\
AgentLM-7B & 24.6 &  -- & 22.3 \\
Lumos-O-7B & \textbf{50.5} & \textbf{65.5} & 24.9 \\
Lumos-I-7B  & 47.1 & 63.6 & \underline{29.4} \\
NAT-7B & \underline{49.1} & \underline{64.4} & \textbf{29.8}\\
\midrule
CodeLlama-13B & \underline{36.1}  & \underline{60.0} & -- \\
AgentLM-13B  & 32.4 & -- & \textbf{29.6} \\
NAT-13B & \textbf{53.8} & \textbf{70.6} & \textbf{29.6} \\
\bottomrule
\end{tabular}}
\caption{Comparison with methods from other papers. We report the best results reported in the corresponding papers.}
\label{tab:others}
\end{table}

\section{Negative-Aware Training}
We first illustrate our motivation for using NAT and then describe our agent framework. We then introduce the whole pipeline of NAT, including data collection, negative-aware reformatting (which is the core part of our method that differentiates it from others), fine-tuning, and inference.
\Cref{fig:illustration} outlines previous methods and our NAT paradigm.
\subsection{Motivation}
Our idea is motivated by two considerations. First, humans learn from mistakes and failures. Failure is often seen as a stepping stone to success, as it offers insights into what does not work and highlights areas that require change or development. We believe that powerful LLMs can also learn these valuable lessons from unsuccessful trajectories.

Second, it is generally hard for humans to compare and learn from experiences when they do not know which experience is successful. Fine-tuning approaches generally treat all examples equally, and LLMs may learn unwanted errors from negative trajectories if negative examples are incorporated directly. Therefore, we add a prefix or suffix to the query to differentiate positive and negative examples, explicitly telling the model whether the following trajectories they learn are correct.

\subsection{Agent Framework}
\paragraph{Prompting Strategy}
As shown in \figref{fig:illustration}, in our agent framework, the process of task resolution is delineated as follows. First, the LLM is provided with a system prompt that outlines (a) the specific task to be addressed (for instance, \texttt{``solve a mathematical problem''}), (b) the tools that are permissible for task execution, and (c) the expected action space and output format (for example, \ex{finish[N]} signifies that \ex{N} is the final answer). We do not provide system prompts in \figref{fig:illustration}, for simplicity.
Second, a query instance is introduced. We prompt the model to answer the query in the \texttt{ReAct} \citep{yao2023react} format, which consists of reasoning texts (referred to as ``thoughts'') and ``actions''. Finally, during the interaction phase, the system executes the LLM-generated actions using the predefined tools, returns the resulting observations back to the LLM, and prompts for subsequent actions until the \textit{finish} action of the task is generated, or the interaction rounds exceed a pre-defined threshold.
Naturally, the task-solving process yields interaction trajectories between the LLM and the environment (i.e., tools in our framework). 

\paragraph{Tools} For math tasks, we design a calculator implemented by SymPy \citep{10.7717/peerj-cs.103}, which takes a math expression as input and outputs the result. For the two question-answering tasks, we design a search tool with the Serper \footnote{https://serper.dev/} API. It takes a search query as input and returns the Google search results. We further re-rank the search results using MPNet \citep{song2020mpnet} and DPR \citep{karpukhin-etal-2020-dense}.

\subsection{The Whole Pipeline}\label{sec:approach}
We introduce the whole pipeline of our negative-aware training paradigm here, where negative-aware reformatting is the core part of the paradigm that enables better agent tuning.

\paragraph{Data Collection} For each task, we obtain the initial questions and corresponding ground truth answers as seed data. We then use GPT-3.5\footnote{We use GPT-3.5-1106 version. Although GPT-4 has the potential to produce even higher quality data, we opted for GPT-3.5 due to cost considerations.}  to generate trajectories three times, each with different temperatures (0.2, 0.5, and 0.7). This allows us to gather a diverse range of positive and negative samples. By comparing predicted answers and ground truth answers, we can label each trajectory as positive or negative.

\paragraph{Negative-Aware Reformatting} Differentiating positive samples from negative samples during the agent tuning process aids in teaching the model to discern between successful and unsuccessful outcomes. We append a string suffix to tell the model whether the training sample is positive or negative. For positive samples, we append \texttt{``Please generate a solution that **correctly** answers the question.''} For negative samples, we append \texttt{``Please generate a solution that **incorrectly** answers the question.''} Unless explicitly stated, we use this in experiments. We also experimented with other reformatting strategies.\footnote{The actual prompts that we use in our experiments are slightly more complex than those provided here.}

\paragraph{Fine-tuning and Inference} We use the reformatted trajectories to fine-tune LLMs. The loss is computed only on the part of the text generated by the LLM, which is similar to fine-tuning a chat model \citep{zheng2023judging}. During inference, we prompt the fine-tuned agent using the prompt for positive examples only.

\begin{table*}[t]
\small
\centering
\resizebox{0.9\textwidth}{!}{
\begin{tabular}{lccccccc}
\toprule
\textbf{Model} & \textbf{\# Positive} & \textbf{Strategy} & \textbf{GSM8K} & \textbf{ASDiv} & \textbf{SVAMP} & \textbf{MultiArith} & \textbf{Average} \\
\midrule
\multirow{3}{*}{LLama-2-7B} & \multirow{3}{*}{2k} & Vanilla  & 35.63 & 60.55 & 47.40 & 80.03 & 55.90 \\
& & NUT & \underline{44.43} & \underline{65.69} & \underline{60.40} & \underline{83.05} & \underline{63.39} \\
& & NAT & \textbf{46.93} & \textbf{66.93} & \textbf{60.80} & \textbf{83.89} & \textbf{64.64} \\
\midrule
\multirow{3}{*}{LLama-2-7B} & \multirow{3}{*}{5k} & Vanilla  & 45.87 & \underline{68.12} & 58.80 & \underline{83.89} & 64.17 \\
& & NUT & \underline{47.54} & 67.03 & \underline{63.50} & 81.71 & \underline{64.95} \\
& & NAT & \textbf{49.05} & \textbf{68.66} & \textbf{64.40} & \textbf{87.58} & \textbf{67.42} \\
\midrule
\multirow{3}{*}{LLama-2-13B} & \multirow{3}{*}{2k} & Vanilla  & 44.43 & 66.49 & 65.40 & \textbf{84.40} & 65.18 \\
& & NUT & \underline{49.43} & \underline{67.72} & \underline{67.60} & 81.37 & \underline{66.53} \\
& & NAT & \textbf{50.64} &\textbf{67.92} & \textbf{68.50} & \underline{83.89} &\textbf{67.74} \\
\midrule
\multirow{3}{*}{LLama-2-13B} & \multirow{3}{*}{5k} & Vanilla & \textbf{54.21} & \textbf{71.28} & 68.30 & \underline{89.26} & \underline{70.76} \\
& & NUT & 51.40 & 70.34 & \underline{68.60} & 86.07 & 69.10 \\
& & NAT & \underline{53.75} & \underline{70.49} & \textbf{70.60} & \textbf{90.27} & \textbf{71.28} \\
\bottomrule
\end{tabular}}
\caption{Overall results for math tasks. Each block is a setting with a specific model and number of positive examples. The best results are \textbf{bolded} and second best results are \underline{underlined}}
\label{tab:math_overall}
\end{table*}

\begin{table}[t]
\small
\centering
\begin{tabular}{ccccc}
\toprule
\multirow{2}{*}{\textbf{Strategy}} & \multicolumn{2}{c}{\textbf{7B}} & \multicolumn{2}{c}{\textbf{13B}} \\
\cmidrule{2-5}
& EM & F1 & EM & F1 \\
\midrule
Vanilla & 27.44 & 36.41 & 27.04 & 36.95 \\
NUT     & 28.04 & 40.96 & 28.24 & \textbf{42.47} \\
NAT     & \textbf{28.80} & \textbf{41.37} & \textbf{28.44} & 42.45 \\
\midrule
NAT-2   & \textbf{29.76} & \textbf{42.51} & \textbf{29.60} & \textbf{43.29} \\
\bottomrule
\end{tabular}
\caption{Results of LLaMA-2-7B and 13B on HotpotQA. All results are reported as the mean score of 5 runs. For the 7B model, we report results using 1,500 negative samples; for the 13B models, we use 2k negative samples. NAT-2 means we divide negative samples into 2 classes based on quality. We discuss this in \Cref{sec:finegrained_nat}.}
\label{tab:hotpot_qa_results}
\end{table}

\begin{table}[t]
\centering
\small
\begin{tabular}{ccc}
\toprule
\textbf{Strategy} & \textbf{7B} & \textbf{13B} \\
\midrule
Vanilla & 55.40  & 53.40 \\
NUT & 62.40  & 61.80 \\
NAT & \textbf{65.80} &  \textbf{64.60}  \\
\bottomrule
\end{tabular}
\caption{Results of LLaMA-2-7B and 13B on StrategyQA with 1000 positive samples and 500 negative samples.}
\label{tab:strategy_qa_results}
\end{table}

\section{Experiments}
\subsection{Experimental Setup}
\paragraph{Datasets} We conduct experiments on mathematical reasoning, multi-hop question answering, and strategic question answering tasks. For math, we collect trajectories using GSM8k \cite{gsm8k} as seed data and test the performance on GSM8k, ASDiv \citep{asdiv}, SVAMP \citep{svamp}, and MultiArith \citep{multiarith}. For question answering, we collect trajectories and test the performance on HotpotQA \cite{yang-etal-2018-hotpotqa} and StrategyQA \cite{geva-etal-2021-aristotle}, respectively. More details are provided in \Cref{app:datasets}.

\paragraph{Baselines} We primarily compare NAT with two baselines. The Vanilla setting uses positive examples only to fine-tune LLMs. This is what previous work \citep{Zeng2023AgentTuningEG,Chen2023FireActTL,Qiao2024AUTOACTAA,Liu2024FromLT} has done. The second setting includes negative examples without adding any prefix or suffix, which we call Negative-Unaware Training (NUT).

\paragraph{Fine-tuning Setup}
We conduct experiments on LLaMA-2-Chat 7B and 13B models \citep{touvron2023llama}.
All the models are fine-tuned for 2 epochs with a batch size of 64. We use a cosine scheduler with 3\% of total steps as the warm-up. The maximum learning rate is set to $5\times10^{-5}$. We train the model with 4$\times$A100 GPUs with DeepSpeed ZeRO 3 stage \citep{Rajbhandari2019ZeROMO}.


\subsection{Results}
\paragraph{Math}
\tabref{tab:math_overall} presents the overall results of the math tasks, from which we observe: 
(1) Incorporating negative examples can improve model performance.
(2) Models with negative-aware training (NAT) not only outperform the corresponding model trained only on positive examples (Vanilla), but also beat the same model trained by directly incorporating negative examples (NUT); and
(3) The improvement of NAT is more substantial when there are fewer positive examples or the model is smaller. Specifically, NAT achieves an $8.74\%$ improvement when using a 7B model with 2k positive examples, and a $0.52\%$ improvement when using a 13B model with 5k positive examples. This highlights the value of NAT in data-scarce scenarios, which is common for agent tuning.

It is worth noting that previous work \citep{Zeng2023AgentTuningEG} has shown that including negative examples harms model performance. We believe this does not contradict our findings: as we discuss in \Cref{sec:analysis_quality,sec:analysis_quantity}, performance is determined by both the quantity and quality of the negative data.

\paragraph{Question Answering}

\Cref{tab:hotpot_qa_results,tab:strategy_qa_results} show the results on HotpotQA and StrategyQA. Here, NAT-2 is a variant of NAT, where we divide negative data into two classes and use different prompts for each, as detailed in \Cref{sec:finegrained_nat}. On HotpotQA, NAT-2 improves performance by more than $2\%$ in EM and $6\%$ in f1 score compared to no negative samples. Compared with NUT, NAT is still about $1\%$  better on EM and f1. On StrategyQA, NAT achieves more than $8\%$ and about $3\%$ improvements compared to no negative samples and NUT, respectively. This suggests that our method is also effective for question-answering tasks.

\section{Analysis}
\label{sec:math_discussion}
\Cref{tab:math_overall,tab:hotpot_qa_results,tab:strategy_qa_results} showcase the capability of LLMs to learn from negative examples when fine-tuned to function as agents. Here, we delve into various factors that could influence the effectiveness of negative-aware training. Specifically, we seek to address the following questions: (1) Given a fixed number of positive examples, how much negative data should be used? (2) What insights does the model gain from negative trajectories? (3) Are all types of negative examples beneficial? and (4) What factors contribute to negative-aware training (NAT) outperforming negative-unaware training (NUT)? Since only the math task contains enough data for our experiments, the analysis is done on the math task.

\subsection{Impact of Training Sample Quantity}
\label{sec:analysis_quantity}
\begin{figure}[t]
    \centering
    \includegraphics[width=\linewidth]{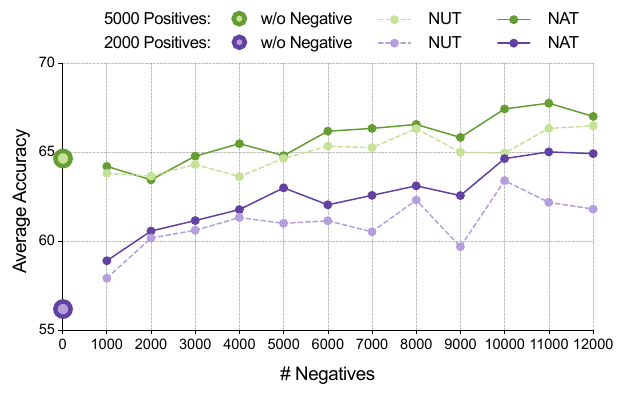}
    \caption{Performance of LLaMA-2-7B for a fixed number of positive samples and variable number of negative samples. }
    \label{fig:math_num_neg}
\end{figure}

\begin{figure}[t] 
    \centering
    \includegraphics[width=\linewidth]{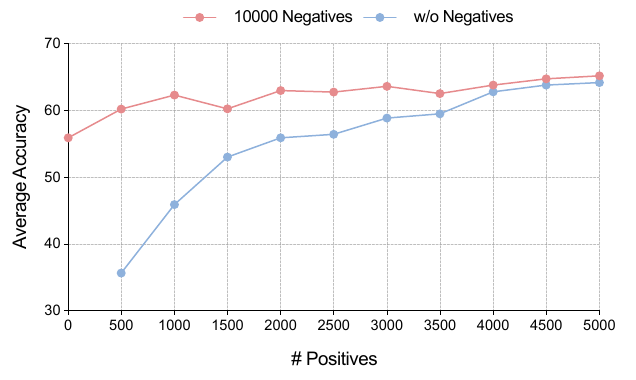}
    \caption{Performance for a fixed number of negative samples (10k) and variable number of positive samples.}
    \label{fig:math_num_pos}
\end{figure}

Our initial analysis focuses on the influence of negative sample quantity. We maintain a constant number of positive samples at 2k and 5k, while adjusting the negative samples from 0 to 12k.  The results, depicted in~\Cref{fig:math_num_neg}, illustrate the relationship between the quantity of negative data and average math task performance. We observe a performance enhancement with an increase in negative data, which plateaus when the volume of negative samples is about 11k in both cases. Due to data availability, we did not experiment with more negatives. 

Based on insights from~\Cref{tab:math_overall} and~\Cref{fig:math_num_neg}, we hypothesize that the ideal ratio of negative samples is not fixed. Instead, it is influenced by two main factors: (1) the number of positive samples, as the improvements are larger for fewer positive samples; 
and (2) the intrinsic quality of the negative samples.

Regarding the first point, we hypothesize that the marginal utility of negative samples diminishes as the quantity of positive samples increases. To test this point, we maintain a constant number of negative samples while varying the quantity of positive samples from 0 to 5k. As depicted in~\Cref{fig:math_num_pos}, there is a diminishing return on the performance added by negative samples as the count of positive samples rises. For the second point, we investigate the effects of negative data quality in~\Cref{sec:analysis_quality}.

\begin{table}[t]
\centering
\resizebox{1\linewidth}{!}{
\begin{tabular}{lccccc}
\toprule
   \textbf{Data} & \textbf{GSM8K} & \textbf{ASDiv} & \textbf{SVAMP} & \textbf{MArith} & \textbf{Avg.}  \\
   \midrule
   \multicolumn{5}{l}{2K positive samples} \\
   \midrule
   Vanilla & 35.63 & 60.55 & 47.40 & 80.03 & 55.90 \\
   NAT-low & 32.98 & 58.72 & 47.60 & 71.64 & 52.74 \\ 
   NAT-high & \textbf{46.93} & \textbf{66.93} & \textbf{60.80} & \textbf{83.89} & \textbf{64.64} \\
   \midrule
   \multicolumn{5}{l}{5K positive samples} \\
   \midrule
   Vanilla & 45.87 & 68.12 & 58.80 & 83.89 & 64.17 \\
   NAT-low & 38.59 & 62.28 & 52.50 & 78.52 & 57.97 \\
   NAT-high & \textbf{49.05} & \textbf{68.66} & \textbf{64.40} & \textbf{87.58} & \textbf{67.42} \\
\bottomrule
\end{tabular}}
\caption{LLaMA-2 7B model results trained with different quality negative data. We use 10k negative samples and experiment with 2k or 5k positive samples.}
\label{tab:math_quality}
\end{table}

\subsection{Data Quality Matters} 
\label{sec:analysis_quality}
We sourced negative data from various models to investigate the impact of negative data quality in NAT. Specifically, we consider the data from GPT-3.5 as high-quality examples. In contrast, we generated 10k negative examples using a fine-tuned LLaMA-2-7B \citep{touvron2023llama} model to represent low-quality data. For experiments, we paired 2k positive examples with 10k negative examples. The outcomes presented in~\Cref{tab:math_quality} underscore the critical role of data quality in NAT. In the 2k positive sample setting, the improvement is $-3.16$ for low quality compared to $+8.74$ for high-quality negative examples. Similarly in the 5k positive sample setting, the improvements are $-6.20$ and $+3.25$, respectively.

\begin{figure}[t]
    \centering
    \includegraphics[width=\linewidth]{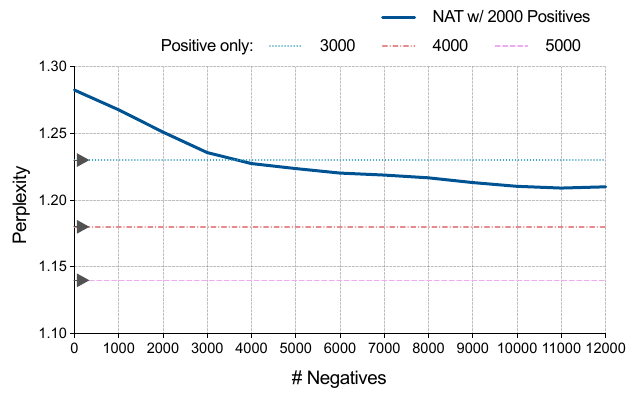}
    \caption{Perplexity for the model trained with 2k positive samples and differing numbers of negative samples. The three dashed lines are perplexity computed on models tuned with differing numbers of positive trajectories (without negatives).}
    \label{fig:math_perplexity}
\end{figure}

\subsection{What does the Model Learn with NAT?}
\paragraph{Learning Reasoning Rather Than Acting} The learnable part in trajectories are thoughts and actions, where thoughts are the reasoning on the current situation and planning for what to do next. Actions are which tool to call and the input to the tool. We analyze the trajectories of the GSM8K \cite{gsm8k} test set generated by LLaMA-2-7B trained with positive examples (Vanilla), NUT, and NAT respectively. \Cref{tab:analysis_learning} shows the accuracy, action error (the percent of incorrectly calling a tool), and average turns (the average number of steps needed to solve a question). Incorporating negative examples also introduces more action errors, which can result in fine-tuned models with more action errors compared to Vanilla. However, after incorporating negative examples, both the accuracy of NUT and NAT increase. This indicates that negative examples mainly work by teaching models with better ``thoughts'' (i.e.\ reasoning and planning). Compared to NUT, NAT achieves significantly fewer action errors and, therefore, better accuracy. This demonstrates that our method works by providing a better trade-off between better ``thoughts'' and more action errors.

\begin{table}[t]
\small
\centering
\begin{tabular}{lccc}
\toprule
\textbf{Strategy} & \textbf{Accuracy} & \textbf{Action Error} & \textbf{\#Avg. Turns}  \\
\midrule
Vanilla & 35.63 & 3.58\% & 3.12\\
NUT & 44.43 & 10.47\% & 3.92 \\
NAT & 46.93 & 7.90\% & 3.71 \\
\bottomrule
\end{tabular}
\caption{Accuracy, action error rate, and number of average turns for models with different training strategies. For Vanilla, the action error in training data is 4.01\%. For NUT and NAT, it is 15.33\%.}
\label{tab:analysis_learning}
\end{table}

\paragraph{Negative Samples Play a Similar Role as Positive Samples} To further explore whether models learn from negative trajectories in the same manner as they learn from positive trajectories, we randomly sample 100 successful trajectories from the training set (as a dev set) and measure the perplexity of models trained with 2k positive examples (not overlapping with the dev set) and varying numbers of negative examples. \Cref{fig:math_perplexity} shows the change in perplexity as the number of negative data increases. The perplexity decreases as more negative data is included, which indicates the model learns to fit successful trajectories with knowledge from failed trajectories. However, this curve seems to be horizontal at the end, and there is still a large gap between the curve with 4k and 5k positives, which shows that some properties or knowledge from successful trajectories can never be learned from failed trajectories.

\begin{table}[t]
\small
\centering
\begin{tabular}{llc}
\toprule
\textbf{Positive} & \textbf{Negative} & \textbf{Average}  \\
\midrule
\texttt{Correct} & \texttt{Incorrect} & 63.55	\\
\texttt{Incorrect} & \texttt{Correct} & 63.33 \\
\texttt{Good} & \texttt{Bad} & 63.91 \\
\midrule
\texttt{A} & \texttt{B} & 63.15 \\
Random string 1 & Random string 2 & 64.04 \\
\bottomrule
\end{tabular}
\caption{Results for models trained on prompts with and without interpretability. Strings in the Positive/Negative column represent prompts (prefixes or suffixes) we use for positive/negative trajectories.}
\label{tab:math_interpretability}
\end{table}

\subsection{Selection of Added Prompts} It has variously been shown that prompts are vital for LLM performance \citep{brown2020language,liu2023pre,Sclar2023QuantifyingLM}. Here, we explore the interpretability of added prompts. More specifically, does the content of the prompt enable LLMs to learn differently from successful and failed trajectories, or simply differentiate these trajectories?
We propose two sets of prompts. One set is prompts with interpretability, such as having the model generate a correct or incorrect trajectory. Another set is prompts without interpretability. For example, different letters can be added as prefixes for queries. \Cref{tab:math_interpretability} shows the results of models trained with interpretable and uninterpretable prompts. Different prompts do not show a large difference in performance, indicating that the performance boost of NAT comes from simply differentiating positive and negative data.


\begin{figure}[t]
    \centering
    \includegraphics[width=\linewidth]{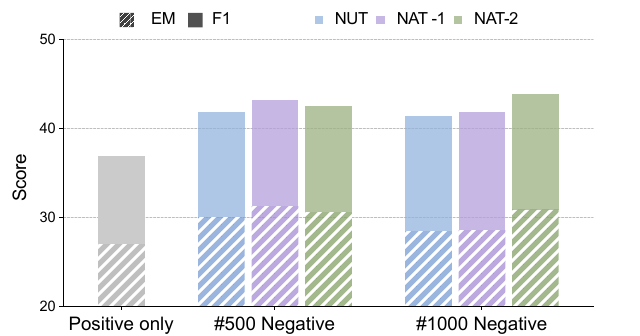}
    \caption{Performance of LLaMA-2-7B on HotpotQA with 500 positive examples and varying numbers of negative examples.}
    \label{fig:qa_num_neg}
\end{figure}

\begin{table}[t]
\centering
\resizebox{1\linewidth}{!}{
\begin{tabular}{cccccc}
\toprule
   \textbf{Strategy} & \textbf{GSM8K} & \textbf{ASDiv} & \textbf{SVAMP} & \textbf{MArith} & \textbf{Avg}  \\
   \midrule
   Vanilla & 29.04 & 55.26 & 45.60 & 80.87 & 52.69 \\ 
   NUT     & 33.50 & \textbf{61.69} & 52.20 & \textbf{86.41} & 58.45 \\ 
   NAT     &  \textbf{36.24} & 61.10 & \textbf{53.90} & 86.24 & \textbf{59.37} \\
   \bottomrule
\end{tabular}}
\caption{LLaMA-2-7B model CoT results fine-tuned using 2k positive samples and 1.6k negative samples.}
\label{tab:math_cot}
\end{table}

\section{Applications} 

In this section, we explore some other applications with our NAT method, showcasing its potential and broad applicability.

\subsection{Fine-grained NAT}
\label{sec:finegrained_nat}
Different negative trajectories contain different degrees of errors. Intuitively, this information also helps models to learn. Therefore, we further propose fine-grained NAT, which divides negative trajectories into different groups based on their quality. During training, different groups will be reformatted with different prompts.
For HotpotQA, in addition to the EM score, each trajectory has an f1 score, measuring the overlap between the predicted and gold answers. We take this as a fine-grained measurement of data quality, where a trajectory with a higher f1 score has better quality. In this way, we can differentiate trajectories based on quality by assigning different prompts. For example, the trajectory is prepended \texttt{``almost wrong''} if its f1 score is smaller than 0.1, and another trajectory is \texttt{``mostly correct''} with an f1 score of 0.9. We denote this NAT with different prompting strategies as NAT-$k$, where $k$ represents how many classes we divide the negative data into based on quality. 
\paragraph{Fine-grained NAT learns more from negative samples}

For NAT-2, we take trajectories with f1 scores equal to 1.0 as positive and assign different prompts for trajectories with f1 scores less than 0.4 and with f1 scores greater than 0.4 less than 1.0. It can be seen from~\Cref{tab:hotpot_qa_results} that the NAT-2 consistently outperforms NAT in all settings, indicating that negative samples associated with finer-grained info are more informative. 
We investigate how the performance changes with differing numbers of negative samples in \Cref{fig:qa_num_neg}. When adding negative samples, the performance increases by a margin, consistent with \Cref{tab:hotpot_qa_results}. However, NAT achieves the best performance with only 500 negative samples, and its performance decreases when adding more negative samples. NAT-2, on the other hand, achieves the best performance with 1k negative samples, consistent with our hypothesis that fine-grained NAT is more beneficial to model training. 

\subsection{Chain-of-Thought Prompting}
\label{sec:cot}
So far we have conducted all experiments on agent scenarios with the \texttt{ReAct} \citep{yao2023react} prompting strategy. In this section, we conduct preliminary experiments to explore whether NAT works well with Chain-of-Thought (CoT) prompting \cite{wei2022chain}. The key difference is that the agent takes an action and receives an observation from the environment iteratively, while CoT generates reasoning steps without taking actions or receiving observations. 

We use GPT-3.5-0125 to generate CoT reasoning steps with three in-context learning \citep{brown2020language} examples on the GSM8k dataset. We then train the model with NAT. \Cref{tab:math_cot} shows the results with CoT prompting. NAT achieves a $6.68\%$ improvement compared to no negative data training (Vanilla). NAT is still about $1\%$ higher compared to directly including negative samples (NUT). The results demonstrate that NAT is also applicable and effective for CoT training, showing its broad applicability.

\section{Conclusion}
In this paper, we first demonstrated that LLMs can learn from failures when fine-tuned as an agent. On the basis of this finding, we propose NAT, a simple and effective method for integrating failed trajectories in fine-tuning agents. We conduct experiments on math and question-answering tasks, and show the superior performance of our method compared to training directly with positive or negative trajectories across tasks and model sizes. 

We are the first to demonstrate the value of negative trajectories in fine-tuning LLMs as agents. Our analysis finds that the quality of negative data is the key to the success of NAT, and models learn similar knowledge to that of positive data (which is much more expensive to attain). NAT is superior because it better utilizes valuable information while restraining learning errors from negative examples.

Negative-aware training is designed to be both agent-agnostic and reasoning strategy-agnostic, making it compatible with various agent strategies, including self-refinement \citep{Madaan2023SelfRefineIR}, reflection \citep{Shinn2023ReflexionLA}, and other agent-tuning frameworks. At the end of this paper, we demonstrated the effectiveness of NAT on Chain-of-Thought (COT) reasoning in mathematical tasks. Moving forward, we aim to assess the applicability and effectiveness of NAT across a broader spectrum of agent frameworks, strategies, and tasks.

\section{Limitations}
Although we have conducted extensive experiments to illustrate the effectiveness of our method, there are still limitations. 
First, similar to previous work in agent tuning, our approach requires the ground truth label of the data, which limits its application. 
Second, we do not experiment with fine-tuning our method on more diverse and powerful models (e.g.\ GPT-3.5) due to time and budget limits.

\bibliography{anthology,custom}

\begin{thebibliography}{40}
\expandafter\ifx\csname natexlab\endcsname\relax\def\natexlab#1{#1}\fi

\bibitem[{Achiam et~al.(2023)Achiam, Adler, Agarwal, Ahmad, Akkaya, Aleman, Almeida, Altenschmidt, Altman, Anadkat et~al.}]{achiam2023gpt}
Josh Achiam, Steven Adler, Sandhini Agarwal, Lama Ahmad, Ilge Akkaya, Florencia~Leoni Aleman, Diogo Almeida, Janko Altenschmidt, Sam Altman, Shyamal Anadkat, et~al. 2023.
\newblock Gpt-4 technical report.
\newblock \emph{arXiv preprint arXiv:2303.08774}.

\bibitem[{Brown et~al.(2020)Brown, Mann, Ryder, Subbiah, Kaplan, Dhariwal, Neelakantan, Shyam, Sastry, Askell et~al.}]{brown2020language}
Tom Brown, Benjamin Mann, Nick Ryder, Melanie Subbiah, Jared~D Kaplan, Prafulla Dhariwal, Arvind Neelakantan, Pranav Shyam, Girish Sastry, Amanda Askell, et~al. 2020.
\newblock Language models are few-shot learners.
\newblock \emph{Advances in neural information processing systems}, 33:1877--1901.

\bibitem[{Chen et~al.(2023)Chen, Shu, Shareghi, Collier, Narasimhan, and Yao}]{Chen2023FireActTL}
Baian Chen, Chang Shu, Ehsan Shareghi, Nigel Collier, Karthik Narasimhan, and Shunyu Yao. 2023.
\newblock \href {https://api.semanticscholar.org/CorpusID:263829338} {Fireact: Toward language agent fine-tuning}.
\newblock \emph{ArXiv}, abs/2310.05915.

\bibitem[{Chen et~al.(2024)Chen, Liu, Wang, Zhang, Liu, Lin, Chen, and Zhao}]{Chen2024AgentFLANDD}
Zehui Chen, Kuikun Liu, Qiuchen Wang, Wenwei Zhang, Jiangning Liu, Dahua Lin, Kai Chen, and Feng Zhao. 2024.
\newblock \href {https://api.semanticscholar.org/CorpusID:268532485} {Agent-flan: Designing data and methods of effective agent tuning for large language models}.
\newblock \emph{ArXiv}, abs/2403.12881.

\bibitem[{Cobbe et~al.(2021)Cobbe, Kosaraju, Bavarian, Chen, Jun, Kaiser, Plappert, Tworek, Hilton, Nakano, Hesse, and Schulman}]{gsm8k}
Karl Cobbe, Vineet Kosaraju, Mohammad Bavarian, Mark Chen, Heewoo Jun, Lukasz Kaiser, Matthias Plappert, Jerry Tworek, Jacob Hilton, Reiichiro Nakano, Christopher Hesse, and John Schulman. 2021.
\newblock Training verifiers to solve math word problems.
\newblock \emph{arXiv preprint arXiv:2110.14168}.

\bibitem[{Geva et~al.(2021)Geva, Khashabi, Segal, Khot, Roth, and Berant}]{geva-etal-2021-aristotle}
Mor Geva, Daniel Khashabi, Elad Segal, Tushar Khot, Dan Roth, and Jonathan Berant. 2021.
\newblock \href {https://doi.org/10.1162/tacl_a_00370} {Did aristotle use a laptop? a question answering benchmark with implicit reasoning strategies}.
\newblock \emph{Transactions of the Association for Computational Linguistics}, 9:346--361.

\bibitem[{Gravitas(2024)}]{Significant_Gravitas_AutoGPT}
Significant Gravitas. 2024.
\newblock \href {https://github.com/Significant-Gravitas/AutoGPT} {Autogpt}.
\newblock \url{https://github.com/Significant-Gravitas/AutoGPT}.

\bibitem[{Green~Jr et~al.(1961)Green~Jr, Wolf, Chomsky, and Laughery}]{green1961baseball}
Bert~F Green~Jr, Alice~K Wolf, Carol Chomsky, and Kenneth Laughery. 1961.
\newblock Baseball: an automatic question-answerer.
\newblock In \emph{Papers presented at the May 9-11, 1961, western joint IRE-AIEE-ACM computer conference}, pages 219--224.

\bibitem[{Karpukhin et~al.(2020)Karpukhin, Oguz, Min, Lewis, Wu, Edunov, Chen, and Yih}]{karpukhin-etal-2020-dense}
Vladimir Karpukhin, Barlas Oguz, Sewon Min, Patrick Lewis, Ledell Wu, Sergey Edunov, Danqi Chen, and Wen-tau Yih. 2020.
\newblock \href {https://doi.org/10.18653/v1/2020.emnlp-main.550} {Dense passage retrieval for open-domain question answering}.
\newblock In \emph{Proceedings of the 2020 Conference on Empirical Methods in Natural Language Processing (EMNLP)}, pages 6769--6781, Online. Association for Computational Linguistics.

\bibitem[{Koncel-Kedziorski et~al.(2016)Koncel-Kedziorski, Roy, Amini, Kushman, and Hajishirzi}]{koncel-kedziorski-etal-2016-mawps}
Rik Koncel-Kedziorski, Subhro Roy, Aida Amini, Nate Kushman, and Hannaneh Hajishirzi. 2016.
\newblock \href {https://doi.org/10.18653/v1/N16-1136} {{MAWPS}: A math word problem repository}.
\newblock In \emph{Proceedings of the 2016 Conference of the North {A}merican Chapter of the Association for Computational Linguistics: Human Language Technologies}, pages 1152--1157, San Diego, California. Association for Computational Linguistics.

\bibitem[{Li et~al.(2023)Li, Yuan, Feng, Pan, Sun, Wang, Wang, and Li}]{Li2023TurningDI}
Yiwei Li, Peiwen Yuan, Shaoxiong Feng, Boyuan Pan, Bin Sun, Xinglin Wang, Heda Wang, and Kan Li. 2023.
\newblock \href {https://api.semanticscholar.org/CorpusID:266375154} {Turning dust into gold: Distilling complex reasoning capabilities from llms by leveraging negative data}.
\newblock \emph{AAAI 2024}.

\bibitem[{Liu et~al.(2024)Liu, Chen, Tian, Zou, Chen, and Cui}]{Liu2024FromLT}
Na~Liu, Liangyu Chen, Xiaoyu Tian, Wei Zou, Kaijiang Chen, and Ming Cui. 2024.
\newblock \href {https://api.semanticscholar.org/CorpusID:266818453} {From llm to conversational agent: A memory enhanced architecture with fine-tuning of large language models}.
\newblock \emph{ArXiv}, abs/2401.02777.

\bibitem[{Liu et~al.(2023{\natexlab{a}})Liu, Yuan, Fu, Jiang, Hayashi, and Neubig}]{liu2023pre}
Pengfei Liu, Weizhe Yuan, Jinlan Fu, Zhengbao Jiang, Hiroaki Hayashi, and Graham Neubig. 2023{\natexlab{a}}.
\newblock Pre-train, prompt, and predict: A systematic survey of prompting methods in natural language processing.
\newblock \emph{ACM Computing Surveys}, 55(9):1--35.

\bibitem[{Liu et~al.(2023{\natexlab{b}})Liu, Yu, Zhang, Xu, Lei, Lai, Gu, Gu, Ding, Men, Yang, Zhang, Deng, Zeng, Du, Zhang, Shen, Zhang, Su, Sun, Huang, Dong, and Tang}]{Liu2023AgentBenchEL}
Xiao Liu, Hao Yu, Hanchen Zhang, Yifan Xu, Xuanyu Lei, Hanyu Lai, Yu~Gu, Yuxian Gu, Hangliang Ding, Kai Men, Kejuan Yang, Shudan Zhang, Xiang Deng, Aohan Zeng, Zhengxiao Du, Chenhui Zhang, Shengqi Shen, Tianjun Zhang, Yu~Su, Huan Sun, Minlie Huang, Yuxiao Dong, and Jie Tang. 2023{\natexlab{b}}.
\newblock \href {https://api.semanticscholar.org/CorpusID:260682249} {Agentbench: Evaluating llms as agents}.
\newblock \emph{ArXiv}, abs/2308.03688.

\bibitem[{Madaan et~al.(2023)Madaan, Tandon, Gupta, Hallinan, Gao, Wiegreffe, Alon, Dziri, Prabhumoye, Yang, Welleck, Majumder, Gupta, Yazdanbakhsh, and Clark}]{Madaan2023SelfRefineIR}
Aman Madaan, Niket Tandon, Prakhar Gupta, Skyler Hallinan, Luyu Gao, Sarah Wiegreffe, Uri Alon, Nouha Dziri, Shrimai Prabhumoye, Yiming Yang, Sean Welleck, Bodhisattwa~Prasad Majumder, Shashank Gupta, Amir Yazdanbakhsh, and Peter Clark. 2023.
\newblock \href {https://api.semanticscholar.org/CorpusID:257900871} {Self-refine: Iterative refinement with self-feedback}.
\newblock \emph{ArXiv}, abs/2303.17651.

\bibitem[{Meurer et~al.(2017)Meurer, Smith, Paprocki, \v{C}ert\'{i}k, Kirpichev, Rocklin, Kumar, Ivanov, Moore, Singh, Rathnayake, Vig, Granger, Muller, Bonazzi, Gupta, Vats, Johansson, Pedregosa, Curry, Terrel, Rou\v{c}ka, Saboo, Fernando, Kulal, Cimrman, and Scopatz}]{10.7717/peerj-cs.103}
Aaron Meurer, Christopher~P. Smith, Mateusz Paprocki, Ond\v{r}ej \v{C}ert\'{i}k, Sergey~B. Kirpichev, Matthew Rocklin, Amit Kumar, Sergiu Ivanov, Jason~K. Moore, Sartaj Singh, Thilina Rathnayake, Sean Vig, Brian~E. Granger, Richard~P. Muller, Francesco Bonazzi, Harsh Gupta, Shivam Vats, Fredrik Johansson, Fabian Pedregosa, Matthew~J. Curry, Andy~R. Terrel, \v{S}t\v{e}p\'{a}n Rou\v{c}ka, Ashutosh Saboo, Isuru Fernando, Sumith Kulal, Robert Cimrman, and Anthony Scopatz. 2017.
\newblock \href {https://doi.org/10.7717/peerj-cs.103} {Sympy: symbolic computing in python}.
\newblock \emph{PeerJ Computer Science}, 3:e103.

\bibitem[{Miao et~al.(2020)Miao, Liang, and Su}]{asdiv}
Shen-yun Miao, Chao-Chun Liang, and Keh-Yih Su. 2020.
\newblock A diverse corpus for evaluating and developing english math word problem solvers.
\newblock In \emph{Proceedings of the 58th Annual Meeting of the Association for Computational Linguistics}, pages 975--984.

\bibitem[{Patel et~al.(2021)Patel, Bhattamishra, and Goyal}]{svamp}
Arkil Patel, Satwik Bhattamishra, and Navin Goyal. 2021.
\newblock \href {https://doi.org/10.18653/v1/2021.naacl-main.168} {Are {NLP} models really able to solve simple math word problems?}
\newblock In \emph{Proceedings of the 2021 Conference of the North American Chapter of the Association for Computational Linguistics: Human Language Technologies}, pages 2080--2094, Online. Association for Computational Linguistics.

\bibitem[{Qiao et~al.(2024)Qiao, Zhang, Fang, Luo, Zhou, Jiang, Lv, and Chen}]{Qiao2024AUTOACTAA}
Shuofei Qiao, Ningyu Zhang, Runnan Fang, Yujie Luo, Wangchunshu Zhou, Yuchen~Eleanor Jiang, Chengfei Lv, and Huajun Chen. 2024.
\newblock \href {https://api.semanticscholar.org/CorpusID:266902590} {Autoact: Automatic agent learning from scratch via self-planning}.
\newblock \emph{ArXiv}, abs/2401.05268.

\bibitem[{Rajbhandari et~al.(2019)Rajbhandari, Rasley, Ruwase, and He}]{Rajbhandari2019ZeROMO}
Samyam Rajbhandari, Jeff Rasley, Olatunji Ruwase, and Yuxiong He. 2019.
\newblock \href {https://api.semanticscholar.org/CorpusID:203736482} {Zero: Memory optimizations toward training trillion parameter models}.
\newblock \emph{SC20: International Conference for High Performance Computing, Networking, Storage and Analysis}, pages 1--16.

\bibitem[{Roy and Roth(2015)}]{multiarith}
Subhro Roy and Dan Roth. 2015.
\newblock \href {https://doi.org/10.18653/V1/D15-1202} {Solving general arithmetic word problems}.
\newblock In \emph{Proceedings of the 2015 Conference on Empirical Methods in Natural Language Processing, {EMNLP} 2015, Lisbon, Portugal, September 17-21, 2015}, pages 1743--1752. The Association for Computational Linguistics.

\bibitem[{Ruan et~al.(2023)Ruan, Dong, Wang, Pitis, Zhou, Ba, Dubois, Maddison, and Hashimoto}]{Ruan2023IdentifyingTR}
Yangjun Ruan, Honghua Dong, Andrew Wang, Silviu Pitis, Yongchao Zhou, Jimmy Ba, Yann Dubois, Chris~J. Maddison, and Tatsunori Hashimoto. 2023.
\newblock \href {https://api.semanticscholar.org/CorpusID:262944419} {Identifying the risks of lm agents with an lm-emulated sandbox}.
\newblock \emph{ArXiv}, abs/2309.15817.

\bibitem[{Sclar et~al.(2024)Sclar, Choi, Tsvetkov, and Suhr}]{Sclar2023QuantifyingLM}
Melanie Sclar, Yejin Choi, Yulia Tsvetkov, and Alane Suhr. 2024.
\newblock \href {https://api.semanticscholar.org/CorpusID:264172710} {Quantifying language models' sensitivity to spurious features in prompt design or: How i learned to start worrying about prompt formatting}.
\newblock \emph{International Conference on Learning Representations}.

\bibitem[{Shinn et~al.(2023)Shinn, Cassano, Labash, Gopinath, Narasimhan, and Yao}]{Shinn2023ReflexionLA}
Noah Shinn, Federico Cassano, Beck Labash, Ashwin Gopinath, Karthik Narasimhan, and Shunyu Yao. 2023.
\newblock Reflexion: Language agents with verbal reinforcement learning.
\newblock \url{https://api.semanticscholar.org/CorpusID:258833055}.

\bibitem[{Song et~al.(2020)Song, Tan, Qin, Lu, and Liu}]{song2020mpnet}
Kaitao Song, Xu~Tan, Tao Qin, Jianfeng Lu, and Tie-Yan Liu. 2020.
\newblock Mpnet: Masked and permuted pre-training for language understanding.
\newblock \emph{arXiv preprint arXiv:2004.09297}.

\bibitem[{Sumers et~al.(2023)Sumers, Yao, Narasimhan, and Griffiths}]{sumers2023cognitive}
Theodore~R Sumers, Shunyu Yao, Karthik Narasimhan, and Thomas~L Griffiths. 2023.
\newblock Cognitive architectures for language agents.
\newblock \emph{arXiv preprint arXiv:2309.02427}.

\bibitem[{Touvron et~al.(2023)Touvron, Martin, Stone, Albert, Almahairi, Babaei, Bashlykov, Batra, Bhargava, Bhosale et~al.}]{touvron2023llama}
Hugo Touvron, Louis Martin, Kevin Stone, Peter Albert, Amjad Almahairi, Yasmine Babaei, Nikolay Bashlykov, Soumya Batra, Prajjwal Bhargava, Shruti Bhosale, et~al. 2023.
\newblock Llama 2: Open foundation and fine-tuned chat models.
\newblock \emph{arXiv preprint arXiv:2307.09288}.

\bibitem[{Wei et~al.(2022)Wei, Wang, Schuurmans, Bosma, Xia, Chi, Le, Zhou et~al.}]{wei2022chain}
Jason Wei, Xuezhi Wang, Dale Schuurmans, Maarten Bosma, Fei Xia, Ed~H Chi, Quoc~V Le, Denny Zhou, et~al. 2022.
\newblock Chain-of-thought prompting elicits reasoning in large language models.
\newblock In \emph{Advances in Neural Information Processing Systems}.

\bibitem[{Weizenbaum(1966)}]{weizenbaum1966eliza}
Joseph Weizenbaum. 1966.
\newblock Eliza—a computer program for the study of natural language communication between man and machine.
\newblock \emph{Communications of the ACM}, 9(1):36--45.

\bibitem[{Wooldridge(1999)}]{wooldridge1999intelligent}
Michael Wooldridge. 1999.
\newblock Intelligent agents.
\newblock \emph{Multiagent systems: A modern approach to distributed artificial intelligence}, 1:27--73.

\bibitem[{Wu et~al.(2023)Wu, Bansal, Zhang, Wu, Zhang, Zhu, Li, Jiang, Zhang, and Wang}]{wu2023autogen}
Qingyun Wu, Gagan Bansal, Jieyu Zhang, Yiran Wu, Shaokun Zhang, Erkang Zhu, Beibin Li, Li~Jiang, Xiaoyun Zhang, and Chi Wang. 2023.
\newblock Autogen: Enabling next-gen llm applications via multi-agent conversation framework.
\newblock \emph{arXiv preprint arXiv:2308.08155}.

\bibitem[{Yang et~al.(2018)Yang, Qi, Zhang, Bengio, Cohen, Salakhutdinov, and Manning}]{yang-etal-2018-hotpotqa}
Zhilin Yang, Peng Qi, Saizheng Zhang, Yoshua Bengio, William Cohen, Ruslan Salakhutdinov, and Christopher~D. Manning. 2018.
\newblock \href {https://doi.org/10.18653/v1/D18-1259} {{H}otpot{QA}: A dataset for diverse, explainable multi-hop question answering}.
\newblock In \emph{Proceedings of the 2018 Conference on Empirical Methods in Natural Language Processing}, pages 2369--2380, Brussels, Belgium. Association for Computational Linguistics.

\bibitem[{Yao et~al.(2023)Yao, Zhao, Yu, Du, Shafran, Narasimhan, and Cao}]{yao2023react}
Shunyu Yao, Jeffrey Zhao, Dian Yu, Nan Du, Izhak Shafran, Karthik Narasimhan, and Yuan Cao. 2023.
\newblock {ReAct}: Synergizing reasoning and acting in language models.
\newblock In \emph{International Conference on Learning Representations (ICLR)}.

\bibitem[{Yin et~al.(2023)Yin, Brahman, Ravichander, Chandu, Chang, Choi, and Lin}]{Yin2023LumosLA}
Da~Yin, Faeze Brahman, Abhilasha Ravichander, Khyathi~Raghavi Chandu, Kai-Wei Chang, Yejin Choi, and Bill~Yuchen Lin. 2023.
\newblock \href {https://api.semanticscholar.org/CorpusID:265128672} {Lumos: Learning agents with unified data, modular design, and open-source llms}.
\newblock \emph{ArXiv}, abs/2311.05657.

\bibitem[{Yoheinakajima(2024)}]{Yoheinakajima2023Babyagi}
Yoheinakajima. 2024.
\newblock \href {https://github.com/Significant-Gravitas/AutoGPT} {Babyagi}.
\newblock \url{https://github.com/Significant-Gravitas/AutoGPT}.

\bibitem[{Zeng et~al.(2023)Zeng, Liu, Lu, Wang, Liu, Dong, and Tang}]{Zeng2023AgentTuningEG}
Aohan Zeng, Mingdao Liu, Rui Lu, Bowen Wang, Xiao Liu, Yuxiao Dong, and Jie Tang. 2023.
\newblock \href {https://api.semanticscholar.org/CorpusID:264306101} {Agenttuning: Enabling generalized agent abilities for llms}.
\newblock \emph{ArXiv}, abs/2310.12823.

\bibitem[{Zhang et~al.(2024)Zhang, Lan, Murthy, Liu, Yao, Tan, Hoang, Yang, Feng, Liu, Awalgaonkar, Niebles, Savarese, Heinecke, Wang, and Xiong}]{Zhang2024AgentOhanaDU}
Jianguo Zhang, Tian Lan, Rithesh Murthy, Zhiwei Liu, Weiran Yao, Juntao Tan, Thai Hoang, Liangwei Yang, Yihao Feng, Zuxin Liu, Tulika Awalgaonkar, Juan~Carlos Niebles, Silvio Savarese, Shelby Heinecke, Huan Wang, and Caiming Xiong. 2024.
\newblock \href {https://api.semanticscholar.org/CorpusID:267897975} {Agentohana: Design unified data and training pipeline for effective agent learning}.
\newblock \emph{ArXiv}, abs/2402.15506.

\bibitem[{Zhao et~al.(2023)Zhao, Huang, Xu, Lin, Liu, and Huang}]{Zhao2023ExpeLLA}
Andrew Zhao, Daniel Huang, Quentin Xu, Matthieu Lin, Y.~Liu, and Gao Huang. 2023.
\newblock \href {https://api.semanticscholar.org/CorpusID:261048772} {Expel: Llm agents are experiential learners}.
\newblock \emph{ArXiv}, abs/2308.10144.

\bibitem[{Zheng et~al.(2023)Zheng, Chiang, Sheng, Zhuang, Wu, Zhuang, Lin, Li, Li, Xing, Zhang, Gonzalez, and Stoica}]{zheng2023judging}
Lianmin Zheng, Wei-Lin Chiang, Ying Sheng, Siyuan Zhuang, Zhanghao Wu, Yonghao Zhuang, Zi~Lin, Zhuohan Li, Dacheng Li, Eric.~P Xing, Hao Zhang, Joseph~E. Gonzalez, and Ion Stoica. 2023.
\newblock \href {http://arxiv.org/abs/2306.05685} {Judging llm-as-a-judge with mt-bench and chatbot arena}.

\bibitem[{Zhou et~al.(2024)Zhou, Zhang, Xiang, Wang, Wu, and Li}]{Zhou2024EnhancingTG}
Qinhao Zhou, Zihan Zhang, Xiang Xiang, Ke~Wang, Yuchuan Wu, and Yongbin Li. 2024.
\newblock \href {https://api.semanticscholar.org/CorpusID:268793648} {Enhancing the general agent capabilities of low-parameter llms through tuning and multi-branch reasoning}.
\newblock \emph{ArXiv}, abs/2403.19962.

\end{thebibliography}

\appendix

\section{Example} \label{app:examples}
\begin{figure*}[t]
    \centering
    \tiny
    \includegraphics[width=1\linewidth]{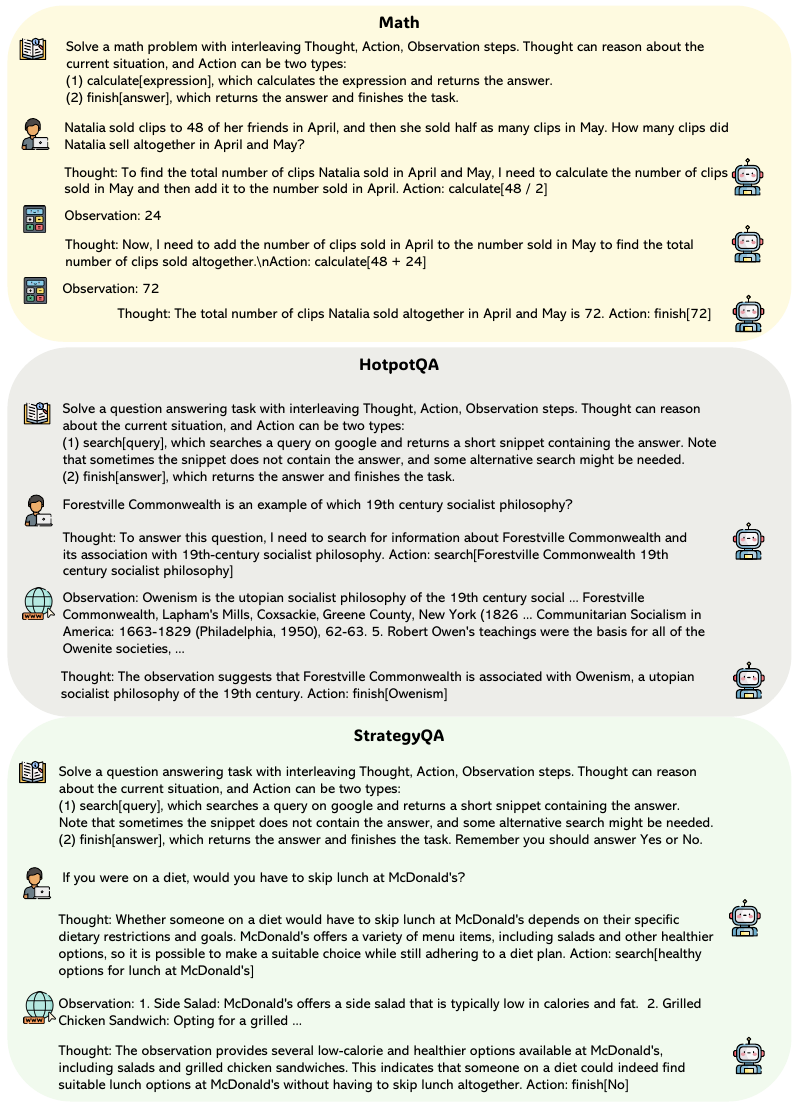}
    \caption{Example trajectories of GSM8k, HotpotQA, and StrategyQA generated by GPT-3.5. }
    \label{fig:full_trajectories}
\end{figure*}

\begin{figure*}[t]
    \centering
    \tiny
    \includegraphics[width=1\linewidth]{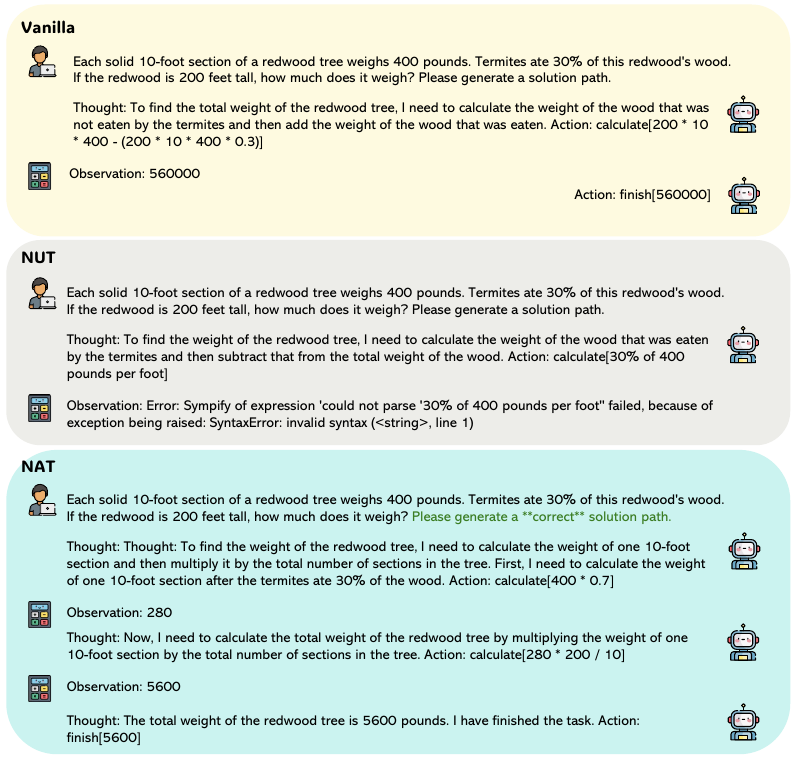}
    \caption{Inference results of models trained with different settings, where Vanilla model gets wrong because of reasoning error, NUT model gets wrong because of action error, and NAT gets correct.}
    \label{fig:trajectories_comparison}
\end{figure*}

\Cref{fig:full_trajectories} shows examples trajectories generated by GPT-3.5. The first turn of each trajectory is the system prompt. \Cref{fig:trajectories_comparison} shows example inference results of models trained with different settings for the same query.

\section{Datasets}
\label{app:datasets}
For mathematical reasoning tasks, we use a dataset of approximately 7k instances from the GSM8K training set as initial seed data, and generate three trajectories with GPT-3.5, as mentioned in \secref{sec:approach}. This process results in a collection of around 9k positive examples and 12k negative examples. Among the positive examples, 5k are unique, indicating that despite multiple attempts, GPT-3.5 fails to solve 2k out of the 7k original questions.

For our experiments, we incorporate 5k unique positive examples from GSM8K to emulate all available positive examples having been generated by GPT-3.5. Additionally, we created a simulated limited dataset using the 2k positive examples generated by ChatGPT. In both scenarios, we include 10k negative examples.

We evaluate different models and training strategies on four test datasets: 
GSM8K \citep{gsm8k}, a high-quality school math word problem dataset containing 1,319 examples (test set), each requiring 2--8 steps to solve;
ASDiv \citep{asdiv}, a math word problem dataset that contains 2,023 examples with diverse language patterns and problem types.
SVAMP \citep{svamp}, a challenge set of math word problems with 1k examples based on perturbing existing datasets \cite{asdiv,koncel-kedziorski-etal-2016-mawps}.
MultiArith \citep{multiarith}, a multi-step arithmetic problem dataset with 596 examples.

For question-answering tasks, we collected trajectories based on HotpotQA \citep{yang-etal-2018-hotpotqa} and StrategyQA \citep{geva-etal-2021-aristotle}. HotpotQA is a Wikipedia-based question-answering dataset where each question requires several steps of reasoning with supporting passages. We use 4k examples from the training set to generate trajectories. StrategyQA is also a multi-step question-answering dataset but the reasoning steps are implicit. The answer to its question is either yes or no. It consists of 2,780 examples, of which 1k is the training set.

Similar to math tasks, we generate three QA trajectories. As discussed in~\Cref{sec:math_discussion}, the quality of negative samples is important for the effectiveness of NAT. For HotpotQA, we filter out trajectories that do not give an answer within a certain number of turns or with a zero f1 score. Finally, we obtain 2k unique positive samples and 2k negative samples. However, we find that 2k examples are enough for performance to saturate and that adding more negative samples causes a performance drop. Therefore, we set the number of HotpotQA positive examples to 500 in the following experiments.

\end{document}